\documentclass[runningheads]{llncs}
\usepackage[table,x11names]{xcolor}
\usepackage{graphicx}
\usepackage{bm} 
\usepackage{wrapfig}

\usepackage{tikz}
\usepackage{comment}
\usepackage{amsmath,amssymb} 
\usepackage{color}

\usepackage[pagebackref,breaklinks,colorlinks,bookmarks=false]{hyperref}
\usepackage{breakcites}
\usepackage{booktabs}
\usepackage{orcidlink}
\definecolor{Gray}{gray}{0.95}

\usepackage[accsupp]{axessibility}  

\definecolor{OliveGreen}{cmyk}{0.64,0,0.95,0.40}
\definecolor{ao(english)}{rgb}{0.0, 0.5, 0.0}
\definecolor{brightmaroon}{rgb}{0.76, 0.13, 0.28}

\begin{document}
\pagestyle{headings}
\mainmatter
\def\ECCVSubNumber{7500}  

\title{Class-Incremental Learning with Cross-Space Clustering and Controlled Transfer} 


\titlerunning{Class-Incremental Learning with CSCCT}
%
\author{Arjun Ashok\inst{1, 2}\orcidlink{0000-0002-2752-2509} \and
K J Joseph\inst{1}\orcidlink{0000-0003-1168-1609} \and
Vineeth N Balasubramanian\inst{1}\orcidlink{0000-0003-2656-0375}}
\authorrunning{A. Ashok et al.}
%
\institute{Indian Institute of Technology Hyderabad\\ \and
PSG College Of Technology, Coimbatore\\ \email{arjun.ashok.psg@gmail.com, \{cs17m18p100001, vineethnb\}@iith.ac.in} \\ \href{https://cscct.github.io}{{https://cscct.github.io}}}
\maketitle

\begin{abstract}
In class-incremental learning, the model is expected to learn new classes continually while maintaining knowledge on previous classes. The challenge here lies in preserving the model's ability to effectively represent prior classes in the feature space, while adapting it to represent incoming new classes. We propose two distillation-based objectives for class incremental learning that leverage the structure of the feature space to maintain accuracy on previous classes, as well as enable learning the new classes. In our first objective, termed {\textbf{cross-space clustering} (CSC)}, we propose to use the feature space structure of the previous model to characterize directions of optimization that maximally preserve the class: directions that all instances of a specific class should collectively optimize towards, and those directions that they should collectively optimize away from. Apart from minimizing forgetting, such a class-level constraint indirectly encourages the model to reliably cluster all instances of a class in the current feature space, and further gives rise to a sense of “herd-immunity”, allowing all samples of a class to jointly combat the model from forgetting the class. Our second objective termed {\textbf{controlled transfer} (CT)} tackles incremental learning from an important and understudied perspective of inter-class transfer. CT explicitly approximates and conditions the current model on the semantic similarities between incrementally arriving classes and prior classes. This allows the model to learn the incoming classes in such a way that it maximizes positive forward transfer from similar prior classes, thus increasing plasticity, and minimizes negative backward transfer on dissimilar prior classes, whereby strengthening stability. We perform extensive experiments on two benchmark datasets, adding our method (CSCCT) on top of three prominent class-incremental learning methods. We observe consistent performance improvement on a variety of experimental settings.

\end{abstract}

\section{Introduction} \label{introduction}

Incremental learning is a paradigm of machine learning where learning objectives are introduced to a model incrementally in the form of \textit{phases} or \textit{tasks}, and the model must dynamically learn new tasks while maintaining knowledge on previously seen tasks. The differences of this setup from a \textit{static training} scenario is that the model is not allowed to \textit{retrain} from scratch on encountering new tasks, and no task information is available upfront.
A fundamental challenge in incremental learning is in the stability-plasticity trade-off \cite{StabPlas}, where stability relates to maintaining accuracy on the previous tasks, while plasticity relates to learning the current task effectively. In their naive form, deep learning models are too plastic; the model changes significantly during training, and incurs \textit{catastrophic forgetting} \cite{FRENCH1999128} of old tasks when exposed to new ones.



Class-incremental learning (CIL) \cite{Rebuffi2017iCaRLIC,masana2021classincremental} is a specific sub-paradigm of incremental learning where tasks are composed of \textit{new classes} and we seek to learn a \textit{unified} model that can \textit{represent and classify} all classes seen so far equally well. The main challenge in class-incremental learning lies in how knowledge over the long stream of classes can be consolidated at every phase. Regularization-based methods \cite{kirkpatrick_ewc,aljundi2018memory,zenkeSI,chaudhry2018riemannian} quantify and preserve important parameters corresponding to prior tasks. Another set of approaches \cite{lopez2017gradient,chaudhry2018efficient,farajtabar2020orthogonal,wang2021training} focus on modifying the learning algorithm to ensure that gradient updates do not conflict with the previous tasks. In dynamic architecture methods \cite{yoon2018lifelong,Yan2021DERDE,Li2019LearnTG,Abati2020ConditionalCG,Rajasegaran2019RandomPS}, the network architecture is modified by expansion or masking when encountering new tasks during learning.
Replay-based methods \cite{Rebuffi2017iCaRLIC,Hou2019LearningAU,Douillard2020PODNetPO,Wu2019LargeSI,Belouadah2019IL2MCI,Castro2018EndtoEndIL,Liu2020MnemonicsTM,Cha2021Co2LCC,lee2019overcoming,Ahn2020SSILSS,Simon2021OnLT,Shin2017ContinualLW,Yin2020DreamingTD} store a subset of each previous task in a separate memory, and replay the tasks when learning a new one, to directly preserve the knowledge on those tasks. 
A wide variety of such replay methods have been developed recently, and have attained promising results in the CIL setting. A number of these methods \cite{Rebuffi2017iCaRLIC,Hou2019LearningAU,Douillard2020PODNetPO,Castro2018EndtoEndIL,Ahn2020SSILSS,Simon2021OnLT,Cha2021Co2LCC,lee2019overcoming} use variants of knowledge distillation \cite{Hinton2015DistillingTK}, where the model and its predictions corresponding to the previous task are utilized to prevent the current task's model from diverging too much from its previous state.

Our work herein falls under distillation-based methods. Prior work has advocated for utilizing distillation to directly constrain an example's position or angle using its position in the previous feature space \cite{Hou2019LearningAU}, to preserve pooled convolutional outputs of an instance \cite{Douillard2020PODNetPO}, or to maintain the distribution of logits that the model's classifier outputs on the data \cite{Ahn2020SSILSS,Rebuffi2017iCaRLIC}. We argue that preserving the features or predictions of a model on independent individual instances are only useful to a certain extent, and do not characterize and preserve properties of a class captured globally by the model as a whole. \textit{Class-level} semantics may be more important to be preserved in the \textit{class-incremental} learning setting, to holistically prevent individual classes from being forgotten.
To this end, we develop an objective termed \textbf{Cross-Space Clustering} (CSC) that characterizes \textit{entire regions} in the feature space that classes should stay away from, and those that a class should belong to, and distills this information to the model. Our objective indirectly establishes multiple goals at once: (i) it encourages the model to cluster all instances of a given class; (ii) ensures that these clusters are well-separated; and (iii) regularizes to preserve \textit{class cluster positions} as a single entity in the feature space. This provides for a class-consolidated distillation objective, prodding instances of a given class to ``unite" and thus prevent the class from being forgotten.

Next, as part of our second objective, we tackle the class-incremental-learning problem from a different perspective. While all prior distillation objectives seek better ways to preserve properties of the learned representations in the feature space \cite{Rebuffi2017iCaRLIC,Hou2019LearningAU,Douillard2020PODNetPO,Castro2018EndtoEndIL,Ahn2020SSILSS,Simon2021OnLT,Cha2021Co2LCC,lee2019overcoming}, we believe that controlling \textit{inter-class transfer} is also critical for class-incremental learning. This comes from the observation that forgetting often results from \textit{negative backward transfer} from new classes to previous classes, and plasticity is ensured when there is \textit{positive forward transfer} from prior classes to new ones \cite{lopez2017gradient}.
To this end, we develop an objective called \textbf{Controlled Transfer} (CT) that controls and regularizes transfer of features between classes at a fine-grained level. We formulate an objective that approximates the relative similarity between an incoming class and all previous classes, and conditions the current task's model on these estimated similarities. This encourages \textit{new classes} to be situated optimally in the feature space, ensuring maximal positive transfer and minimal negative transfer. 


A unique characteristic of our objectives is their ability to extend and enhance existing distillation-based CIL methodologies, without any change to their methodologies. We verify this by adding our objectives to three prominent and state-of-the art CIL methods that employ distillation in their formulation: iCARL \cite{Rebuffi2017iCaRLIC}, LUCIR \cite{Hou2019LearningAU} and PODNet \cite{Douillard2020PODNetPO}. We conduct thorough experimental evaluations on benchmark incremental versions of large-scale datasets like CIFAR-100 and ImageNet subset. We perform a comprehensive evaluation of our method, considering a wide variety of experimental settings. We show that our method consistently improves incremental learning performance across datasets and methods, at no additional cost. We further analyze and present ablation studies on our method, highlighting the contribution of each of our components.

\section{Related Work} \label{related}

\subsection{Incremental Learning} 

In an incremental setting, a model is required to consistently learn new tasks, without compromising performance on old tasks. Incremental learning methodologies can be split into five major categories, each of which we review below.

\textbf{Regularization-based} methods focus on quantifying the importance of each parameter in the network, to prevent the network from excessively changing the important parameters pertaining to a task. These methods include EWC \cite{kirkpatrick_ewc}, SI \cite{zenkeSI}, MAS \cite{aljundi2018memory} and RWalk \cite{chaudhry2018riemannian}. A recent method ELI \cite{joseph2022energy}, introduces an energy based implicit regularizer which helps to  alleviate forgetting.

\textbf{Algorithm-based} methods seek to avoid forgetting from the perspective of the network training algorithm. They modify gradients such that updates to weights do not not deteriorate performance on previously seen tasks. Methods such as GEM \cite{lopez2017gradient}, A-GEM \cite{chaudhry2018efficient}, OGD \cite{farajtabar2020orthogonal} and NSCL \cite{wang2021training} fall under this category. Meta-learning based methods \cite{javed2019meta,kj2020meta} are also useful while learning continually. 

\textbf{Architecture-based} methods modify the network architecture dynamically to fit more tasks, by expanding the model by adding more weights \cite{yoon2018lifelong,Yan2021DERDE}, or masking and allocating subnetworks to specific tasks \cite{Serr2018OvercomingCF}, or by gating the parameters dynamically using a task identifier \cite{Abati2020ConditionalCG}.

\textbf{Exemplar-based} methods (also called replay-based or rehearsal methods) assume that a small subset of every class can be stored in a memory at every phase. They replay the seen classes later along with the incoming new classes, directly preventing them from being forgetten. One set of works focus on reducing the recency bias due to the new classes being in majority at every phase \cite{Wu2019LargeSI,Belouadah2019IL2MCI,Hou2019LearningAU,Castro2018EndtoEndIL}. Another set of works focus on optimizing which samples to choose as exemplars to better represent the class distributions \cite{Liu2020MnemonicsTM,Aljundi2019GradientBS}.

\textbf{Distillation-based} methods use the model learned until the previous task as a teacher and provide extra supervision to the model learning the current tasks (the student). Since the data of the previous tasks are inaccessible, these methods typically enforce distillation objectives on the current data \cite{Li2018LearningWF,Dhar2019LearningWM}, data from an exemplar memory \cite{Hou2019LearningAU,Douillard2020PODNetPO,Rebuffi2017iCaRLIC,Castro2018EndtoEndIL,Ahn2020SSILSS,kj2021incremental,joseph2021towards}, external data \cite{lee2019overcoming} or synthetic data \cite{Zhai2019LifelongGC}. Since our method falls under this category, we extend our discussion on related methods below.

Early works in this category distill logit scores \cite{Li2018LearningWF,Rebuffi2017iCaRLIC} or attention maps \cite{Dhar2019LearningWM} of the previous model. iCARL \cite{Rebuffi2017iCaRLIC} proposes to enforce distillation on new tasks as well exemplars from old tasks, along with herding selection of exemplars and nearest-mean-of-exemplars (NME) based classification. GD \cite{lee2019overcoming} calibrates the confidence of the model's outputs using external unlabelled data, and propose to distill the calibrated outputs instead. LUCIR \cite{Hou2019LearningAU} introduces a less-forget constraint that encourages the orientation of a sample in the current feature space to be similar to its orientation in the old feature space. Apart from that, LUCIR proposes to use cosine-similarity based classifiers and a margin ranking loss that mines hard negatives from the new classes to better separate the old class, additionally avoiding ambiguities between old and new classes. PODNet \cite{Douillard2020PODNetPO} preserves an example's representation throughout the model with a spatial distillation loss. 
SS-IL \cite{Ahn2020SSILSS} show that general KD preserves the recency bias in the distillation, and propose to use task-wise KD. Co2L \cite{Cha2021Co2LCC} introduces a contrastive learning based self-supervised distillation loss that preserves the exact feature relations of a sample with its augmentations and other samples from the dataset. 
GeoDL \cite{Simon2021OnLT} introduces a term that enhances knowledge distillation by performing KD across low-dimensions path between the subspaces of the two models, considering the gradual shift between models.

The main difference of our cross-space clustering objective from these works is that we do not optimize to preserve the properties of individual examples, and instead preserve the previously learned semantics or properties of each \textit{class} in a \textit{holistic manner}. 
Our formulation takes into account the global position of a class in the feature space, and optimizes all samples of the class towards the same region, making the model indifferent to instance-level semantics.
Further, classes are supervised with specific ``negative" regions all over the feature space, also intrinsically giving rise to better separation between class clusters.

Our controlled transfer objective, on the other hand, attempts to regularize transfer between tasks. MER \cite{riemer2018learning}, an algorithm-based method is related to our high-level objective. MER works in the online continual learning setup, combining meta-learning \cite{finn2017model, nichol2018first} with replay. Their method optimizes such that the model receives weight updates are restricted to those directions that agree with the gradients of prior tasks. 
Our objective proposes to utilize the \textit{structure} of the feature space of the previous model to align the current feature space, in order to maximize transfer.
Our novelty here lies in how we explicitly approximate \textit{inter-class semantic similarities} in a \textit{continual task stream}, and utilize them to appropriately position new tasks representations, regularizing transfer.

\subsection{Knowledge Distillation}

Hinton et al. \cite{Hinton2015DistillingTK} introduced knowledge distillation (KD) in their work as a way to transfer knowledge from an ensemble of teacher networks to a smaller student network. They show that there is dark knowledge in the logits of a trained network that can give more structure about the data, and use them as soft targets to train the student. Since then, a number of other works have explored variants of KD. Attention Transfer \cite{Zagoruyko2017PayingMA} focused on the attention maps of the network instead of the logits, while FitNets \cite{Romero2015FitNetsHF} also deal with intermediate activation maps of a network. Several other papers have enforced criteria based on multi-layer representations of a network \cite{Yim2017AGF,Huang2017LikeWY,Kim2018ParaphrasingCN,Ahn2019VariationalID,Koratana2019LITLI}. 

Among these, our controlled transfer objective shares similarities with a few works that propose to exploit the mutual relation between data samples for distillation. Tung and Mori \cite{Tung2019SimilarityPreservingKD} propose a distillation objective that enforces an L2 loss in the student that constraints the similarities between activation maps of two examples to be consistent with those of the teacher. 
Relational KD \cite{Park2019RelationalKD} additionally propose to preserve the angle formed by the three examples in the feature space by means of a triplet distillation objective. Extending this direction, Correlation Congruence \cite{Peng2019CorrelationCF} models relations between examples through higher-order kernels, to enforce the same objectives with better relation estimates.

The difference of our controlled transfer objective from these works lies in the high-level objective in the context of the incremental learning setting, as well as the low-level formulation in terms of the loss objective. All the above works propose to use sample relations in the feature space to provide additional supervision to a student model 
by regularizing the feature relations of the student.
The main challenge in incremental learning is how we can reduce the effect that a new class has on the representation space, to minimize forgetting.

Our objective also exploits sample relations in the feature space, however our novelty lies in how we estimate a measure of {relative similarity} between an \textit{unseen class} and each previously seen class, and utilize them to control where the \textit{new samples} are located in the embedding space, in relation to the old samples. 
Our specific formulation indirectly promotes forward transfer of features from prior classes similar to the new class, and simultaneously prevents negative backward transfer of features from the new class to dissimilar previous classes
.

\section{Method} \label{method-section}

We briefly introduce the problem setting in
Sec.~\ref{problem}. Next, we explain in detail our learning objectives in Sec.~\ref{obj1} and Sec.~\ref{obj2} respectively, and discuss the overall objective function in Sec.~\ref{obj3}.

\subsection{Problem Setting} \label{problem}

In the incremental learning paradigm, a set of tasks $\mathcal{T}_t=\{\tau_1, \tau_2,\cdots,\tau_t\}$ is introduced to the model over time, where $\mathcal{T}_t$ represents the tasks that the model has seen until time step $t$. $\tau_t$ denotes the task introduced at time step $t$, which is composed of images and labels sampled from its corresponding data distribution: $\tau_t = (\mathbf{x}^{\tau_{t}}_i, y^{\tau_{t}}_i) \sim p_{data}^{\tau_t}$. Each task $\tau_t$ contains instances from a disjoint set of classes. $\mathcal{F}^{\mathcal{T}_t}$ denotes the model at time step $t$
. 
Without loss of generality, $\mathcal{F}^{\mathcal{T}_t}$ can be expressed as a composition of two functions: $\mathcal{F}^{\mathcal{T}_t}(\mathbf{x})=(\mathcal{F}_{\bm \phi}^{\mathcal{T}_t} \circ \mathcal{F}_{\bm \theta}^{\mathcal{T}_t})(\mathbf{x})$, where $\mathcal{F}_{\bm \phi}^{\mathcal{T}_t}$ represents a feature extractor, and  $\mathcal{F}_{\bm \theta}^{\mathcal{T}_t}$ denotes a classifier. The challenge in incremental learning is to learn a model that can represent and classify all seen classes equally well, at any point in the task stream.

While training the model $\mathcal{F}^{\mathcal{T}_t}$ on the task $\tau_t$, the model does not have access to all the data from previous tasks. Exemplar-based methods \cite{Rebuffi2017iCaRLIC, Aljundi2019GradientBS, Wu2019LargeSI, Liu2020MnemonicsTM, Belouadah2019IL2MCI} sample a very small subset of each task data $e_{t} \in \tau_{t}$ at the end of every task $\tau_{t}$ and store it in a memory buffer $\mathcal{M}_{t} = \{e_1, e_2,\cdots,e_t\}$, which contains a subset of data from all tasks seen until time $t$. When learning a new task at time step $t$, the task's data $\tau_{t}$ is combined with samples from the memory containing exemplars of each previous task $\mathcal{M}_{t-1}$. Therefore, the dataset that the model has at time step $t$ is $\mathcal{D}_t=\tau_{t} \cup \mathcal{M}_{t-1}$. In distillation-based methods, we assume access to the previous model $\mathcal{F}^{\mathcal{T}_{t-1}}$ which has learned the stream of tasks $\mathcal{T}_{t-1}$. The model $\mathcal{F}^{\mathcal{T}_{t-1}}$ is frozen and not updated, and is instead used to guide the learning of the current model $\mathcal{F}^{\mathcal{T}_{t}}$. 
Effectively utilising the previous model is key to balancing stability and plasticity. Excess constraints tied to the model can prevent the current task from being learned, and poor constraints can lead to easy forgetting of previous tasks.

\begin{figure}[t!]
\centering
\includegraphics[width=1\textwidth, height=0.25\textheight]{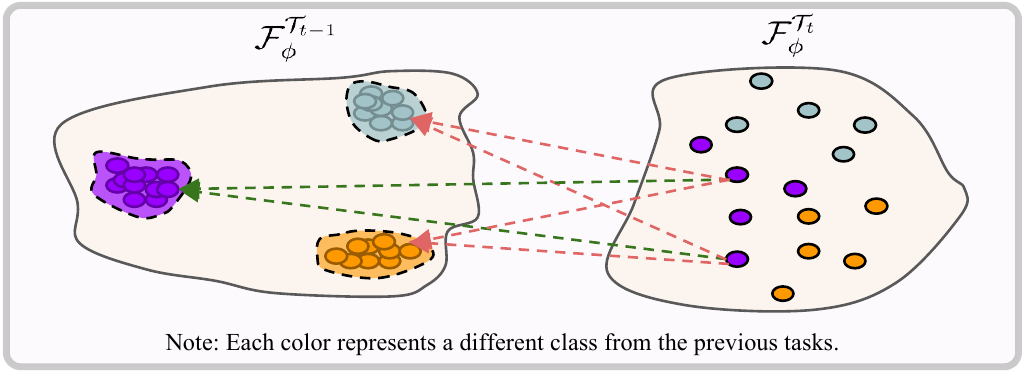}
\caption{We illustrate our {cross-space clustering} (CSC) objective. We show instances from 3 classes from the stream $\mathcal{T}_{t-1}$, and their positions in $\mathcal{F}^{\mathcal{T}_{t-1}}_{\phi}$ and $\mathcal{F}^{\mathcal{T}_{t}}_{\phi}$ respectively. 
Classes are well represented in $\mathcal{F}^{\mathcal{T}_{t-1}}_{\phi}$, however their representations are dispersed in the $\mathcal{F}^{\mathcal{T}_{t}}_{\phi}$. 
Here we illustrate the constraint imposed on an instance of the \textcolor{violet}{violet} class, based on the cluster position of its own class (indicated by the \textcolor{green!45!black}{green} arrows) and the positions of every other class (indicated by the \textcolor{red}{red} arrows). Note how the exact same constraint is applied on all instances of a class (illustrated here with $2$ instances of the \textcolor{violet}{violet} class). 
Best viewed in color.
}
\label{fig:csc}
\end{figure}

\subsection{Cross-Space Clustering} \label{obj1}

Our \textit{cross-space clustering (CSC)} objective alleviates forgetting by distilling class-level semantics and inducing tight clusters in the feature space. CSC leverages points across the entire feature space of the previous model $\mathcal{F}^{\mathcal{T}_{t-1}}$, to identify regions that a class is optimized to stay within, and other harmful regions that it is prevented from drifting towards. 
We illustrate our cross-space clustering objective in Fig.~\ref{fig:csc}. 

Consider that the model $\mathcal{F}^{\mathcal{T}_{t}}$ is trained on mini-batches $\mathcal{B}$ = $\{x_i, y_i\}_{i=1}^{k}$ sampled from $D_t$. 
Our cross-space clustering objective enforces the following loss on the model:




\begin{equation}
     L_{Cross-Cluster} = \frac{1}{k^2} \sum_{i=1}^{k} \sum_{j=1}^{k} 
    \left(1 - 
    \cos (\mathcal{F}_{\bm \phi}^{\mathcal{T}_t} (x_i),
        \mathcal{F}_{\bm \phi}^{\mathcal{T}_{t-1}} (x_j))
    \right) * ind(y_i == y_j)
    \label{eqn:obj1}
\end{equation}

\noindent where $ind$ is an indicator function that returns \textbf{1} when its inputs are equal and \textbf{-1} otherwise\footnote{Note how this is different from a typical indicator function that returns 0 when the inputs are not equal.} , and $\cos (a,b)$ denotes the cosine similarity between two vectors $a$ and $b$.

\textbf{Physical Interpretation}:
For pairs of samples $x_i$ and $x_j$, $\mathcal{F}^{\mathcal{T}_{t}}(x_i)$ is enforced to minimize cosine distance with $\mathcal{F}^{\mathcal{T}_{t-1}}(x_j)$ when they are of the same class ($y_i == y_j$), and maximize cosine distance with $\mathcal{F}^{\mathcal{T}_{t-1}}(x_j)$ when they are of different classes ($y_i != y_j$). We expand upon the objective and its implications separately below.

\textbf{Explanation}:
Consider that there are $l$ examples of class $n$ in the considered batch of size $k$, and hence $k-l$ samples belonging to classes other than $n$.

Sample $x_i$ from class $n$ is allowed to see the previous feature positions of all of the $l$ samples of the same class in $\mathcal{B}$, and is regularized to be \textit{equally close}
to all these positions. Since multiple positions are used in the previous feature space and equal constraints are applied, points only see an approximation of its class (cluster) in the previous feature space, and do not see individual feature positions. 
This inherently removes the dependency of the distillation on the specific position of the sample within its class, and instead optimizes the sample to move towards a point that can preserve the class as a whole.

Next, \textit{every sample} $x_i$ belonging to a class $n$ in the batch is given the \textit{exact same constraints} with no difference. In our case, all samples belonging to a class are optimized towards the mean of the class's embeddings in the previous space. This leads to all of them being optimized \textit{jointly} to a single stark region belonging to their class. Repeating this process for several iterations implicitly leads to model to implicitly \textit{cluster} all samples of a class in the \textit{current} feature space $\mathcal{F}^{\mathcal{T}_{t}}$ in the specific characterized regions. With respect to clustering, an important point is that our loss is \textit{cross-space} in the sense, it does not encourage clustering of features of a class using features from the same model \cite{Cha2021Co2LCC}, that would neither exploits prior knowledge about the inter-class distances, nor imposes any constraints on the location of the classes. Our formulation instead encourages a model to keep all these clusters at specific points provided by the previous feature space, thereby directly distilling and preserving the \textit{cluster positions} in the feature space as well. Hence, our objective uses approximate cluster positions from $\mathcal{F}^{\mathcal{T}_{t-1}}$
to in-turn cluster samples at specific positions in $\mathcal{F}^{\mathcal{T}_{t}}$. 
Since all samples are optimized towards the same region, all points of the class are optimized to \textit{unite} and jointly protect and preserve the class. Such a formulation gives rise to a sense of \textit{herd-immunity} of classes from forgetting, which better preserves the classes as the model drifts.

Finally, with very few exemplars stored per-class in the memory, our objective proposes to maximally utilize the memory\footnote{A batch of sufficient size typically contains at least one sample from each previous class, serving as a rough approximation of the memory} as well as the current task, leveraging them to identify \textit{negative regions} that an instance is maintained to lie away from. 
Through multiple iterations of optimization, $x_i$ belonging to class $n$ is enforced to stay equally away from the positions of all other $k-l$ examples from the entire previous space. This indirectly tightens the cluster of class $n$ in $\mathcal{F}^{\mathcal{T}_{t}}$ along multiple directions in the feature space. 

\textbf{Differences from prior work}:
Prior distillation-based methods \cite{Hou2019LearningAU, Douillard2020PODNetPO, Simon2021OnLT} only apply \textit{sample-to-sample} cross-space constraints, to preserve the representational properties of the previous space. 
The core difference of our method from all others lies in how it applies \textit{class-to-region} constraints. Here, \textit{class} denotes how all samples of a class are jointly optimized with the same constraints, and \textit{region} denotes how the samples are optimized towards and away from specific \textit{regions} instead of towards individual points.

\begin{figure}[t!]
\centering
\includegraphics[width=0.9\textwidth, height=0.3\textheight]{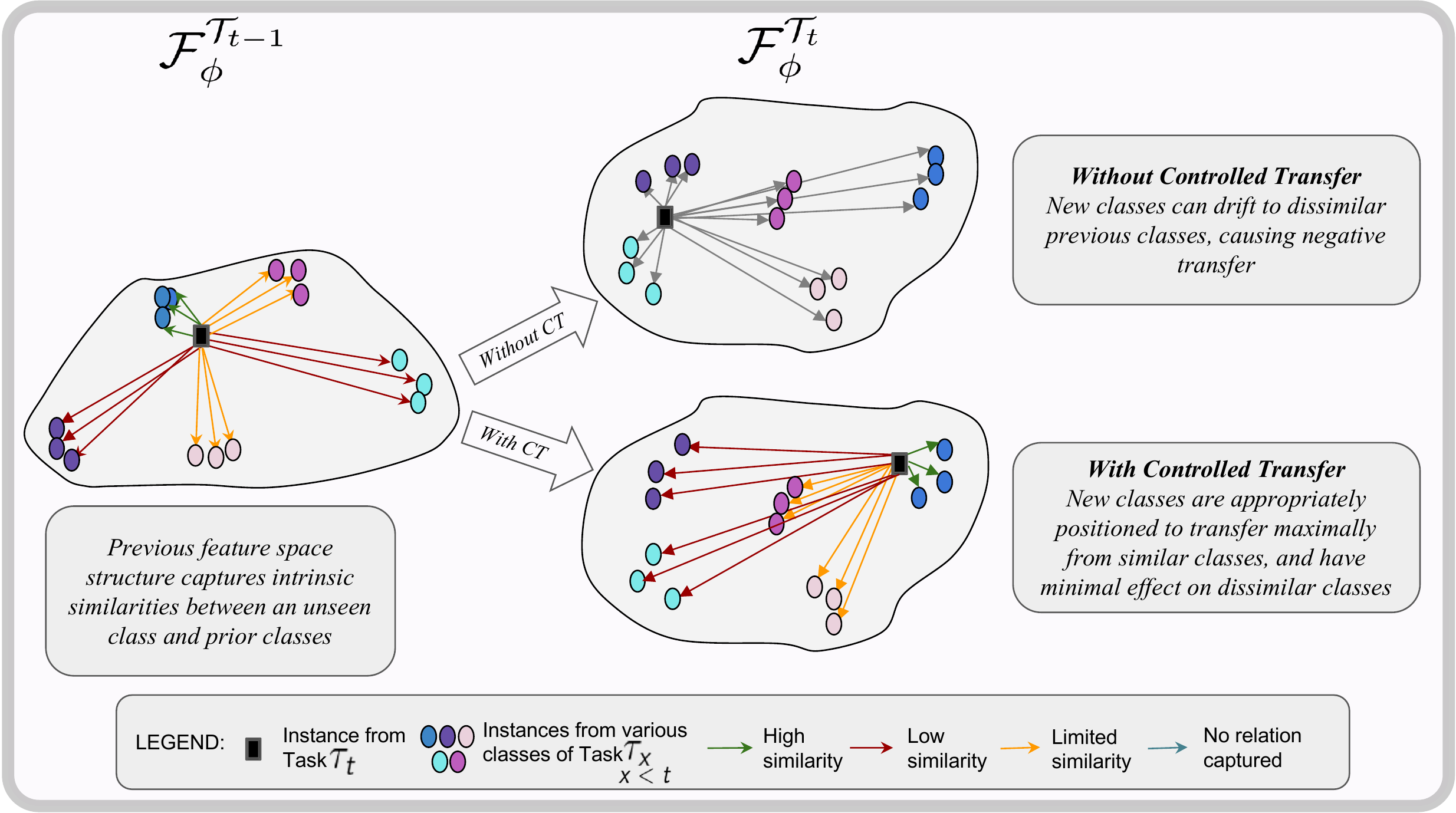}
\caption{We illustrate our \textbf{controlled transfer} objective. We show the positions of instances from five random classes taken from previous tasks $\tau_x; x<t$, and one unseen incoming class from the current task $\tau_t$, in $\mathcal{F}^{\mathcal{T}_{t-1}}_{\phi}$ and $\mathcal{F}^{\mathcal{T}_{t}}_{\phi}$ respectively. 
With our objective, the new task instances are regularized to position themselves appropriately, to prevent negative transfer to dissimilar classes, and to encourage positive transfer from similar classes (best viewed in color)
}
\label{fig:ct}
\end{figure}

\subsection{Controlled Transfer} \label{obj2}

Catastrophic forgetting has been previously characterized to arise 
due to a variety of reasons - the inability to access enough data of previous tasks \cite{Wu2019LargeSI,Hou2019LearningAU,Belouadah2019IL2MCI}, change in important parameters \cite{kirkpatrick_ewc,zenkeSI,aljundi2018memory}, representation drift \cite{Li2018LearningWF,Douillard2020PODNetPO,Cha2021Co2LCC}, conflicting gradients \cite{lopez2017gradient, chaudhry2018efficient, farajtabar2020orthogonal, wang2021training} and insufficient capacity of models \cite{yoon2018lifelong,Yan2021DERDE,Abati2020ConditionalCG}. 
However, all these works ignore the semantic similarities between tasks and their relation to forgetting. 
We argue that knowing the degree of semantic similarity between two classes can, in fact, be very useful in incremental learning: When a previous class is \textit{dissimilar} to the class currently being learned, the model must learn to treat the previous class distinctively and minimally impact it, so that the semantic specialities of that class are not erased. Conversely, when there is a previous class which is \textit{similar} to the class currently being learned, the model must maximally transfer features from that class, to learn the current class in the best possible way. With these goals, we propose an incremental learning objective that  \textit{explicitly quantifies} inter-class similarities, and \textit{controls} transfer between classes in every phase. Fig.~\ref{fig:ct} illustrates our controlled transfer objective.

\textbf{Notation}: We first describe the general notation that we use to denote the similarity between samples in a space. Consider two samples $x_i$ and $x_j$ from a dataset $D_k$, and a model $\mathcal{F}^{\mathcal{T}_k}$. We denote the similarity between $x_i$ and $x_j$ computed on the feature space of $\mathcal{F}^{\mathcal{T}_k}$ as $z^{\mathcal{T}_k}_{x_i,x_j} = \cos(\mathcal{F}_{\bm \phi}^{\mathcal{T}_k} (x_i), \mathcal{F}_{\bm \phi}^{\mathcal{T}_k} (x_j))$ where $\cos(a,b)$ denotes cosine similarity between two vectors $a$ and $b$. 
We denote the the normalized distribution of similarities that an individual sample $x_i$ has with \textit{every sample} in $D_k$, in the feature space of $\mathcal{F}_{}^{\mathcal{T}_k}$ as 
\begin{equation}
    H^{\mathcal{T}_k}_{x_i,D_k,T} = \left\{\frac{(z^{\mathcal{T}_k}_{x_i,x_j}/T)}
                 {\sum_{g=1}^{|D_k|}(z^{\mathcal{T}_k}_{x_i,x_g}/T)}
      \right\}_{j=1}^{|D_k|}
\end{equation}

where $T$ is the temperature that is used to control the entropy of the distribution. 
$H^{\mathcal{T}_k}_{x_i,D_k,T}$ is a row matrix, where the value in each column $j$ of the matrix denotes the normalized similarity between $x_i$ and $x_j$, relative to every sample in the dataset $D_k$. 


\textbf{Formulation}: We first aim to estimate the similarities between a \textit{new class} $\mathcal{C}_{new} \in \tau_t$ and every previously seen class $\mathcal{C}_{old} \in \tau_k \in \mathcal{T}_{t-1}$. 
$\mathcal{C}_{old}$ is well represented the model $\mathcal{F}^{\mathcal{T}_{t-1}}$; the new class $\mathcal{C}_{new}$ has not yet been learned by any model. 
It is not possible to use the drifting feature space of $\mathcal{F}^{\mathcal{T}_{t}}$ to represent $\mathcal{C}_{new}$; even representing $\mathcal{C}_{new}$ once it has been learned by $\mathcal{F}^{\mathcal{T}_{t}}$ would heavily bias the representations towards $\mathcal{C}_{new}$  
due to the well-known recency bias \cite{Wu2019LargeSI, Ahn2020SSILSS}. 
Our formulation instead proposes to utilize the \textit{dark knowledge} that the \textit{previous model} possesses about an \textit{unseen class}:
if the representations of an unseen class $\mathcal{C}_{new}$ lie relatively close to the class representations of a previous class in $\mathcal{F}^{\mathcal{T}_{t-1}}$, it indicates that the two classes share semantic similarities. On the other hand, if the representations of an unseen class $\mathcal{C}_{new}$ lie relatively far from a previous class in $\mathcal{F}^{\mathcal{T}_{t-1}}$, it indicates that the two classes do not have any semantic features in common. 
Note how the similarities for $\mathcal{C}_{new}$ given by $\mathcal{F}^{\mathcal{T}_{t-1}}$ are \textit{unbiased} as the model has never seen $\mathcal{C}_{new}$, and hence can be used as approximations to the semantic similarities.
We propose to use these approximate similarities captured by $\mathcal{F}^{\mathcal{T}_{t-1}}$ in our objective explained below. 


Consider a mini-batch of $B^{\mathcal{T}_{t}}_n$ of size $s$ that contains samples $\{(x^{\mathcal{T}_t}_{i}, y^{\mathcal{T}_t}_{i})\}$ randomly sampled from $D_t$. This mini-batch $B^{\mathcal{T}_{t}}_n$ is composed of $p$ samples from the current task denoted by $P = {(x^{\tau_t}_{i}, y^{\tau_t}_{i})}_{i=1}^{p}$, and $q$ samples taken from the memory, denoted by $Q = {(x^{\tau_{k}}_{i}, y^{\tau_k}_{i})}_{i=1}^{q}$, where $k < t$. 
In an effort to control the transfer between a new and an old sample, our objective regularizes the normalized similarity (closeness) that a sample from the current task $(x^{\tau_t}_{i}, y^{\tau_t}_{i}) \in \tau_t$ has with every sample from any previous class $(x^{\tau_k}_{i}, y^{\tau_k}_{i})$, where $k < t$. 
This is enforced by minimizing the KL Divergence of the similarity distribution of $x^{\tau_t}_{i} \in P$ over $Q$, in the \textit{current space} $\mathcal{F}^{\mathcal{T}_t}_{\phi}$,
with the similarity distribution computed in the \textit{previous space} $\mathcal{F}^{\mathcal{T}_{t-1}}_{\phi}$,
 as follows 
\begin{equation}
     L_{Transfer} = \frac{1}{p} \sum_{i=1}^{p} KL(H^{\tau_t}_{x_i,Q,T} || H^{\tau_{t-1}}_{x_i,Q,T})
    \label{eqn:obj2}
\end{equation}

This loss modifies the position of the current classes in the current feature space $\mathcal{F}^{\mathcal{T}_t}_{\phi}$ such that they have \textit{high similarity} with (lie close to) prior classes that are \textit{very similar}.
This encourages \textit{positive forward transfer} of features to the current classes from selected previous classes that are similar, as both their embeddings are optimized to have high similarity in the current space. 
This helps the model learn the current task better by leveraging transferred features, and lessens the impact that the new task has on the representation space.
Conversely, as the embeddings are optimized to have \textit{low similarity} with (lie far from) those previous classes that are \textit{dissimilar} to it, it discourages (negative) backward transfer from the current classes to those dissimilar classes.
Consequently, the features of these specific classes are further shielded from being erased, leading to the semantics of those classes being preserved more in the current space, which directly results in lesser forgetting of those classes.





\subsection{Final Objective Function} \label{obj3}

The independent nature of our objectives make them suitable to be applied on top of any existing method to improve its performance.
Our final objective combines $L_{Cross-Cluster}$ (\ref{eqn:obj1}) and $L_{Transfer}$ (\ref{eqn:obj2}) with appropriate coefficients:

\begin{equation}
  L_{CSCCT} = L_{method} + \alpha*L_{Cross-Cluster} + \beta*L_{Transfer}  
  \label{eqn:final}
\end{equation}

where $L_{method}$ denotes the objective function of the base method used, and $\alpha$ and $\beta$ are loss coefficients for each of our objectives respectively. We term our method \textbf{CSCCT}, indicating \textbf{C}ross-\textbf{S}pace \textbf{C}lustering and \textbf{C}ontrolled \textbf{T}ransfer.


\begin{table*}[t]
\centering\setlength{\tabcolsep}{2pt}
\caption{The table shows results on \textbf{CIFAR100} when our method is added to three top-performing approaches \cite{Rebuffi2017iCaRLIC, Hou2019LearningAU, Douillard2020PODNetPO}. The \textcolor{red}{red} subscript highlights the relative improvement. $\mathcal{B}$ denotes the number of classes in the first task. $\mathcal{C}$ denotes the number of classes in every subsequent task.}
\label{tab:cifar100_results}
\resizebox{\textwidth}{!}{%
\setlength{\tabcolsep}{7pt} 
\begin{tabular}{l|lll|lll}
\toprule
\multicolumn{1}{c|}{Dataset} & \multicolumn{6}{c}{CIFAR100} \\ \midrule
\multicolumn{1}{c|}{Settings} & \multicolumn{3}{c|}{$\mathcal{B}=50$} & \multicolumn{3}{c}{$\mathcal{B}=\mathcal{C}$} \\ \midrule
\multicolumn{1}{c|}{Methods} & \multicolumn{1}{c}{$\mathcal{C}=1$} & \multicolumn{1}{c}{$\mathcal{C}=2$} & \multicolumn{1}{c|}{$\mathcal{C}=5$} & \multicolumn{1}{c}{$\mathcal{C}=1$} & \multicolumn{1}{c}{$\mathcal{C}=2$} & \multicolumn{1}{c}{$\mathcal{C}=5$} \\ \midrule
\multicolumn{1}{l|}{iCaRL \cite{Rebuffi2017iCaRLIC}} & $43.39$ & $48.31$ & \multicolumn{1}{l|}{$54.42$} & $30.92$ & $36.80$ & $44.19$ \\
\rowcolor{Gray}
\multicolumn{1}{l|}{iCaRL + CSCCT} & $46.15_{\color{red}\textbf{+2.76}}$ & $51.62_{\color{red}\textbf{+3.31}}$ & \multicolumn{1}{l|}{$56.75_{\color{red}\textbf{+2.33}}$} & ${{34.02}_{\color{red}\textbf{+3.1}}}$ & ${{\textbf{39.60}}_{\color{red}\textbf{+2.8}}}$ & ${{46.45}_{\color{red}\textbf{+2.26}}}$ \\
\multicolumn{1}{l|}{LUCIR \cite{Hou2019LearningAU}} & $50.26$ & $55.38$ & \multicolumn{1}{l|}{$59.40$} & $25.40$ & $31.93$ & $42.28$ \\
\rowcolor{Gray}
\multicolumn{1}{l|}{LUCIR + CSCCT} & $52.95_{\color{red}\textbf{+2.69}}$ & ${56.49}_{\color{red}\textbf{+1.13}}$ & \multicolumn{1}{l|}{${{62.01}_{\color{red}\textbf{+2.61}}}$} & ${28.12}_{\color{red}\textbf{+2.72}}$ & ${34.96}_{\color{red}\textbf{+3.03}}$ & $44.03_{\color{red}\textbf{+1.55}}$ \\
\multicolumn{1}{l|}{PODNet \cite{Douillard2020PODNetPO}} & $56.88$ & $59.98$ & \multicolumn{1}{l|}{$62.66$} & $33.58$ & $36.68$ & $45.27$ \\
\rowcolor{Gray}
\multicolumn{1}{l|}{PODNet + CSCCT} & $\textbf{58.80}_{\color{red}\textbf{+1.92}}$ & $\textbf{61.10}_{\color{red}\textbf{+1.12}}$ & \multicolumn{1}{l|}{$\textbf{63.72}_{\color{red}\textbf{+1.06}}$} & $\textbf{36.23}_{\color{red}\textbf{+2.65}}$ 
& ${{39.3}}_{\color{red}\textbf{+2.62}}$ & ${\textbf{47.8}}_{\color{red}\textbf{+2.53}}$ \\ \bottomrule
\end{tabular}%
}
\end{table*}

\section{Experiments and Results} \label{exps}

We conduct extensive experiments adding our method to three prominent methods in class-incremental learning \cite{Rebuffi2017iCaRLIC, Hou2019LearningAU, Douillard2020PODNetPO}. 

\textbf{Protocols:} Prior work has experimented with two CIL protocols: \textbf{a)} training with half the total number of classes in the first task, and equal number of classes in each subsequent task \cite{Hou2019LearningAU, Douillard2020PODNetPO, Wu2019LargeSI}, and \textbf{b)} training with the same number of classes in each task, including the first \cite{Rebuffi2017iCaRLIC, Ahn2020SSILSS, Castro2018EndtoEndIL}. The first setting has the advantage of gaining access to strong features in the first task, while the second tests an extreme continual learning setting; both these are plausible in a real-world incremental classification scenario. We experiment with both these protocols to demonstrate the applicability of our method. On CIFAR100, classes are grouped into $1$, $2$ and $5$ classes per task. On ImageNet-Subset, the classes are split into $2$, $5$ and $10$ classes per task.
Hence, we experiment on both \textit{long streams of} \textit{small tasks}, as well as \textit{short streams of} \textit{large tasks}. 

\textbf{Datasets and Evaluation Metric:}
Following prior works \cite{Hou2019LearningAU, Douillard2020PODNetPO, Rebuffi2017iCaRLIC, Castro2018EndtoEndIL}, we test on the incremental versions of CIFAR-100 \cite{krizhevsky2009cifar} and ImageNet-Subset \cite{Rebuffi2017iCaRLIC}. CIFAR100 contains $100$ classes, with $500$ images per class, and each of dimension $32\times32$. ImageNet-Subset is a subset of the ImageNet-1k dataset \cite{deng2009imagenet}, and contains $100$ classes, with over $1300$ images per class. Each image is of size $224\times224$.
All our results denote average incremental accuracy
. 
We follow the original papers in their inference methodology: On LUCIR \cite{Hou2019LearningAU} and PODNet \cite{Douillard2020PODNetPO}, classification is performed as usual using the trained classifier, while on iCARL \cite{Rebuffi2017iCaRLIC}, classification is based on nearest-mean-of-exemplars.

\section{Implementation Details}
Following prior work \cite{Hou2019LearningAU, Douillard2020PODNetPO}, we use a ResNet-32 and ResNet-18 on CIFAR100 and ImageNet-Subset respectively.
On CIFAR100, we use a batch size of $128$ and train for $160$ epochs, with an initial learning rate of $4e^{-1}$ that is decayed by $0.1$ at the $80^{th}$ and $120^{th}$ epochs respectively. On ImageNet-Subset, we use a batch size of $64$ and train for $90$ epochs, with an initial learning rate of $2e^{-2}$ that is decayed by $0.1$ at the $30^{th}$ and $60^{th}$ epochs respectively.
We use \textit{herding selection} \cite{Rebuffi2017iCaRLIC} for exemplar sampling, and an exemplar memory size of $20$.


\begin{table*}[t]
\centering\setlength{\tabcolsep}{2pt}
\caption{The table shows results on \textbf{ImageNet-Subset} when our method is added to three top-performing approaches \cite{Rebuffi2017iCaRLIC, Hou2019LearningAU, Douillard2020PODNetPO}. The \textcolor{red}{red} subscript highlights the relative improvement. $\mathcal{B}$ denotes the number of classes in the first task. $\mathcal{C}$ denotes the number of classes in every subsequent task.}
\label{tab:imagenet_subset_results}
\resizebox{\textwidth}{!}{%
\setlength{\tabcolsep}{7pt} 
\begin{tabular}{l|lll|lll}
\toprule
\multicolumn{1}{c|}{Dataset} & \multicolumn{6}{c}{ImageNet-Subset} \\ \midrule
\multicolumn{1}{c|}{Settings} & \multicolumn{3}{c|}{$\mathcal{B}=50$} & \multicolumn{3}{c}{$\mathcal{B}=\mathcal{C}$} \\ \midrule
\multicolumn{1}{c|}{Methods} & \multicolumn{1}{c}{$\mathcal{C}=2$} & \multicolumn{1}{c}{$\mathcal{C}=5$} & \multicolumn{1}{c|}{$\mathcal{C}=10$} & \multicolumn{1}{c}{$\mathcal{C}=2$} & \multicolumn{1}{c}{$\mathcal{C}=5$} & \multicolumn{1}{c}{$\mathcal{C}=10$} \\ \midrule
\multicolumn{1}{l|}{iCaRL \cite{Rebuffi2017iCaRLIC}} & $55.81$ & $57.34$ & \multicolumn{1}{l|}{$65.97$} & $40.75$ & $55.92$ & $60.93$ \\
\rowcolor{Gray}
\multicolumn{1}{l|}{iCaRL + CSCCT} & ${{57.01}_{\color{red}\textbf{+1.2}}}$ & ${{58.37}_{\color{red}\textbf{+1.03}}}$ & \multicolumn{1}{l|}{${{66.82}_{\color{red}\textbf{+0.8}}}$} & ${{42.46}_{\color{red}\textbf{+1.71}}}$ & ${{57.45}_{\color{red}\textbf{+1.53}}}$ & ${{62.60}_{\color{red}\textbf{+1.67}}}$ \\
\multicolumn{1}{l|}{LUCIR \cite{Hou2019LearningAU}} & $60.44$ & $66.55$ & \multicolumn{1}{l|}{$70.18$} & $36.84$ & $46.40$ & $56.78$ \\
\rowcolor{Gray}
\multicolumn{1}{l|}{LUCIR + CSCCT} & ${61.52}_{\color{red}\textbf{+1.08}}$ & ${67.91}_{\color{red}\textbf{+1.36}}$ & \multicolumn{1}{l|}{${71.33}_{\color{red}\textbf{+1.15}}$} & ${37.86}_{\color{red}\textbf{+1.02}}$ & ${47.55}_{\color{red}\textbf{+1.15}}$ & $58.07_{\color{red}\textbf{+1.29}}$ \\
\multicolumn{1}{l|}{PODNet \cite{Douillard2020PODNetPO}} & $67.27$ & $73.01$ & \multicolumn{1}{l|}{$75.32$} & $44.94$ & $58.23$ & $66.24$ \\
\rowcolor{Gray}
\multicolumn{1}{l|}{PODNet + CSCCT} & $\textbf{68.91}_{\color{red}\textbf{+1.64}}$ & $\textbf{74.35}_{\color{red}\textbf{+1.34}}$ & \multicolumn{1}{l|}{$\textbf{76.41}_{\color{red}\textbf{+1.09}}$} & $\textbf{46.06}_{\color{red}\textbf{+1.12}}$ & $\textbf{59.43}_{\color{red}\textbf{+1.2}}$ & ${\textbf{67.49}}_{\color{red}\textbf{+1.25}}$ \\ \bottomrule
\end{tabular}%
}
\end{table*}

\subsection{Quantitative Results}

We add our method to three state-of-the-art class-incremental learning methodologies: iCARL \cite{Rebuffi2017iCaRLIC}, LUCIR \cite{Hou2019LearningAU} and PODNet \cite{Douillard2020PODNetPO}.  Table \ref{tab:cifar100_results} showcases results on CIFAR100, and Table \ref{tab:imagenet_subset_results} showcases results on ImageNet-Subset. We see a consistent improvement across all settings and methods when CSCCT is added to them. Specifically, on CIFAR100, adding CSCCT to iCARL \cite{Rebuffi2017iCaRLIC}, LUCIR \cite{Hou2019LearningAU} and PODNet \cite{Douillard2020PODNetPO} provides strong relative improvement of $2.76\%$, $2.28\%$ and $1.99\%$ respectively averaged across all settings, while on the much more high-dimensional ImageNet-Subset, adding our method to the respective baselines provides consistent relative improvements of $1.32\%$, $1.17\%$ and $1.35\%$.

Evaluating iCARL \cite{Rebuffi2017iCaRLIC}, LUCIR \cite{Hou2019LearningAU} and PODNet \cite{Douillard2020PODNetPO} on the equal class protocol show that LUCIR \cite{Hou2019LearningAU} suffers from a severe performance degradation due to its inherent reliance on a large initial task, while iCARL \cite{Rebuffi2017iCaRLIC} and PODNet \cite{Douillard2020PODNetPO} do not. 
On CIFAR100, simply adding our method to iCARL \cite{Rebuffi2017iCaRLIC} gives it strong boosts of $2.2\%-3.1\%$ in this setting, bringing it much closer to the state-of-the-art PODNet \cite{Douillard2020PODNetPO}. 
Overall, our method improves performance consistently across both settings, showing that our formulation \textit{does not rely on a large initial task to learn strong representations.}

\section{Ablation Study and Analysis} \label{ablation}

\begin{figure}[t!]
    \centering
    \includegraphics[width=.45\textwidth, height=0.2\textheight]{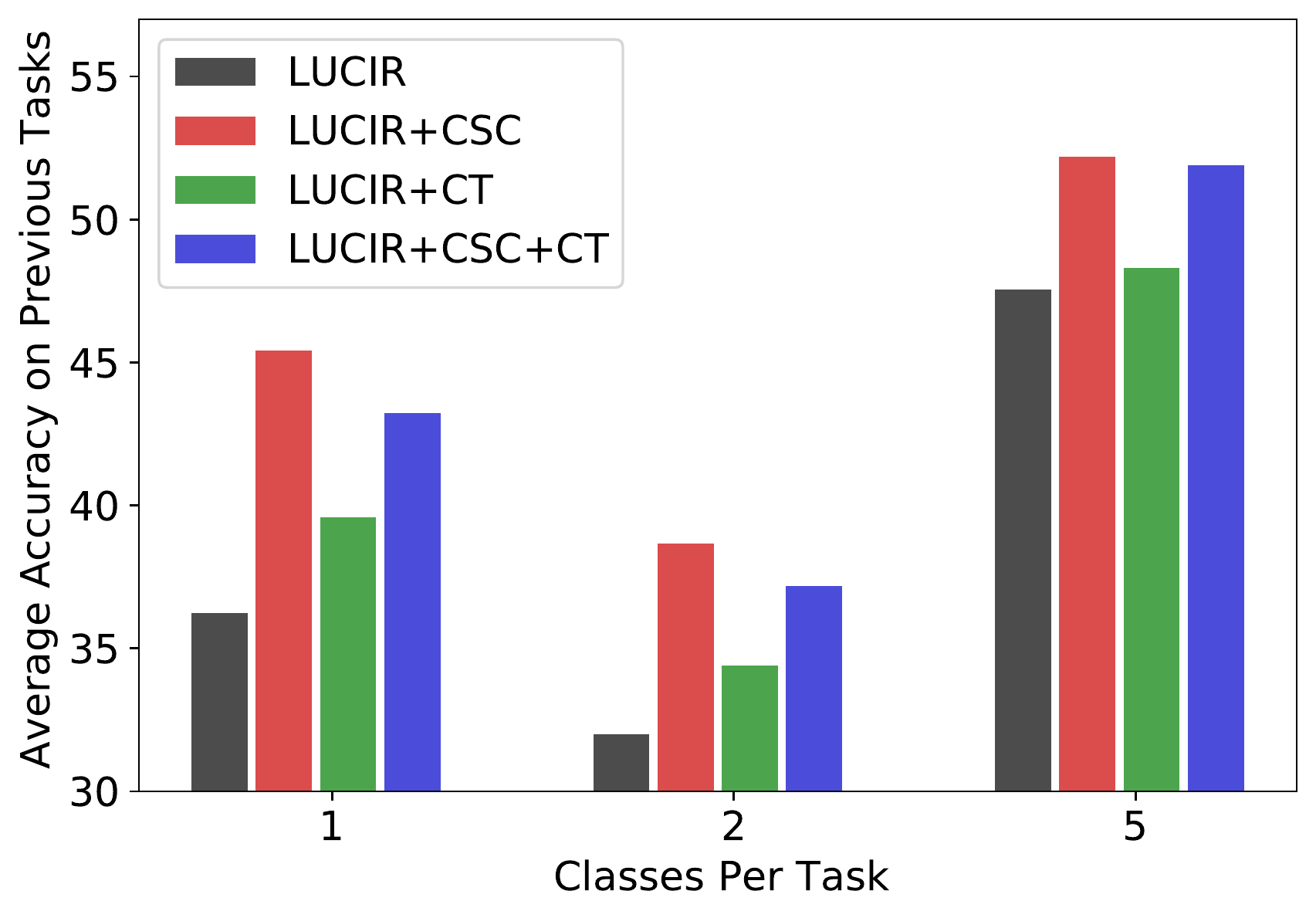}\hfill
    \includegraphics[width=.45\textwidth, height=0.2\textheight]{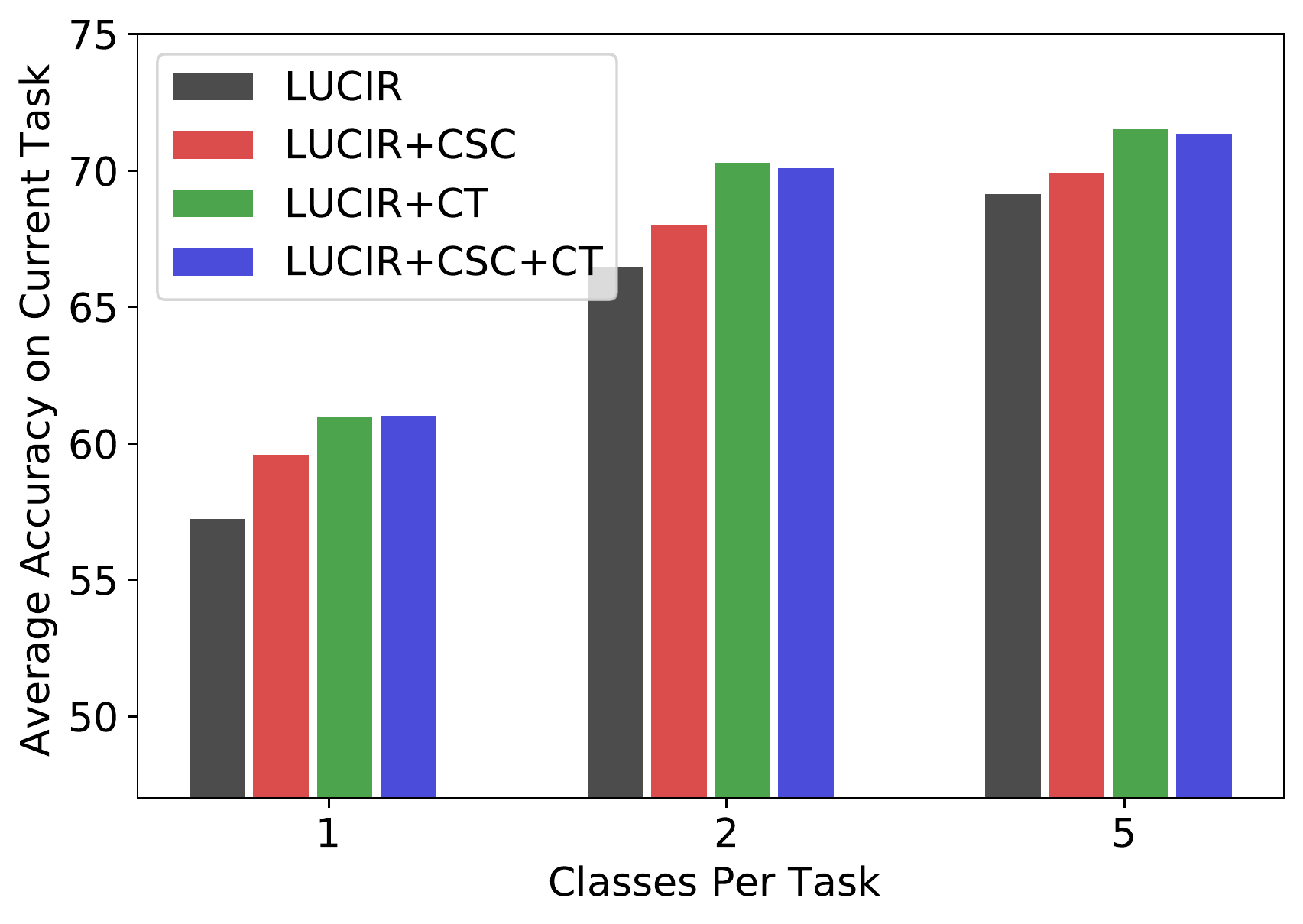}\hfill
    \caption{Average accuracy on previous tasks (APT) and average accuracy on the current task (ACT), plotted across various settings on the CIFAR-100 dataset}
    \label{fig:barplots}
\end{figure}

\subsection{Effect of Each Component on Average Incremental Accuracy}

\setlength\intextsep{0pt} 
\begin{wraptable}{r}{0.6\textwidth}
{\scriptsize
\caption{Ablating each objective on CIFAR100. \textcolor{brightmaroon}{Maroon} denotes $2^{nd}$ best result.}
\label{table:ablation_table1}}
\begin{center}
\scriptsize
\begin{tabular}{l|lll|lll}
    \toprule
    \multicolumn{1}{c|}{Settings} & 
    \multicolumn{3}{c|}{$\mathcal{B}=50$} & \multicolumn{3}{c}{$\mathcal{B}=\mathcal{C}$} 
    \\ \midrule
    \multicolumn{1}{c|}{Methods} & 
    \multicolumn{1}{c}{$\mathcal{C}=1$} & \multicolumn{1}{c}{$\mathcal{C}=2$} & \multicolumn{1}{c|}{$\mathcal{C}=5$} & \multicolumn{1}{c}{$\mathcal{C}=1$} & \multicolumn{1}{c}{$\mathcal{C}=2$} & \multicolumn{1}{c}{$\mathcal{C}=5$}   
    \\ \midrule
    \multicolumn{1}{l|}{LUCIR \cite{Hou2019LearningAU}} & 
    $50.26$ & $55.38$ & \multicolumn{1}{l|}{$59.4$} & 
    $25.4$ & $31.93$ & $42.28$ \\
    \multicolumn{1}{l|}{LUCIR + CSC} & 
    $\textcolor{brightmaroon}{52.04}$ & $\textcolor{brightmaroon}{55.95}$ & \multicolumn{1}{l|}{$60.45$} & 
    $\textcolor{brightmaroon}{27.16}$ & $32.89$ & $42.98$ \\
    \multicolumn{1}{l|}{LUCIR + CT} &
    $51.5$ & $55.87$ & \multicolumn{1}{l|}{$\textcolor{brightmaroon}{61.97}$} & 
    $26.53$ & $\textcolor{brightmaroon}{33.98}$ & $\textcolor{brightmaroon}{43.69}$ \\
    \multicolumn{1}{l|}{LUCIR + CSCCT} & 
    $\textbf{52.95}$ & $\textbf{56.49}$ & \multicolumn{1}{l|}{$\textbf{62.01}$} & 
    $\textbf{28.12}$ & $\textbf{34.96}$ & $\textbf{44.03}$ \\
    \bottomrule
\end{tabular}
\end{center}
\end{wraptable}

In Table~\ref{table:ablation_table1}, we ablate each component of our objective.
Each of our objectives can improve accuracy independently. 
In particular, CSC is more effective when the number of classes per task is extremely low, while CT stands out in the improvement it offers when there are more classes per task. Overall, combining our objectives achieves the best performance across all settings. 


\subsection{Effect of Each Component on Stability/Plasticity}

To further investigate how each component is useful specifically in the  incremental learning setup, we look into how each component improves the stability and plasticity of the model under various settings.
The left plot of Fig.~\ref{fig:barplots} shows the average accuracy on previous tasks (denotes as APT).
This serves as an indicator of the \textbf{stability} of the model. Mathematically, APT can be expressed as

\begin{equation}
    APT = \frac{
        \sum_{t=2}^{T}{
            \left(\frac{
                \sum_{k=1}^{t-1}{
                    Acc(\mathcal{F}^{\mathcal{T}_{t}}, \tau_{k})
                }
            }
            {t-1}\right)
        }
    }
    {T-1}
\end{equation}

where $Acc(\mathcal{F}^{\mathcal{T}_{t}}, \tau_k)$ denotes accuracy of model $\mathcal{F}^{\mathcal{T}_{t}}$ on the test set of task $k$.

The right plot of Fig.~\ref{fig:barplots} shows the average accuracy on the current task (denoted as ACT).
This serves as an indicator of the \textbf{plasticity} of the model.  ACT is expressed as
\begin{equation}
    ACT = \frac{
        \sum_{t=1}^{T}{
            Acc(\mathcal{F}^{\mathcal{T}_{t}}, \tau_{t})
        }
    }
    {T}
\end{equation}

Across all considered settings, \textit{both of our objectives} increase \textit{stability} as well as \textit{plasticity} of the base model. However, one can see that the effect of the \textbf{CSC objective} is much more pronounced on the \textbf{stability} of the model. This aligns with intuition that the CSC helps in {preserving previous classes better} in the representation space. At the same time, the \textbf{CT objective} impacts the \textbf{plasticity} consistently more than the CSC objective, as it mainly aims at {appropriately positioning the current task samples} to maximize transfer. 

\subsection{Embedding Space Visualization}


\begin{figure}[t!]
    \centering
    \includegraphics[width=.45\textwidth, height=0.2\textheight]{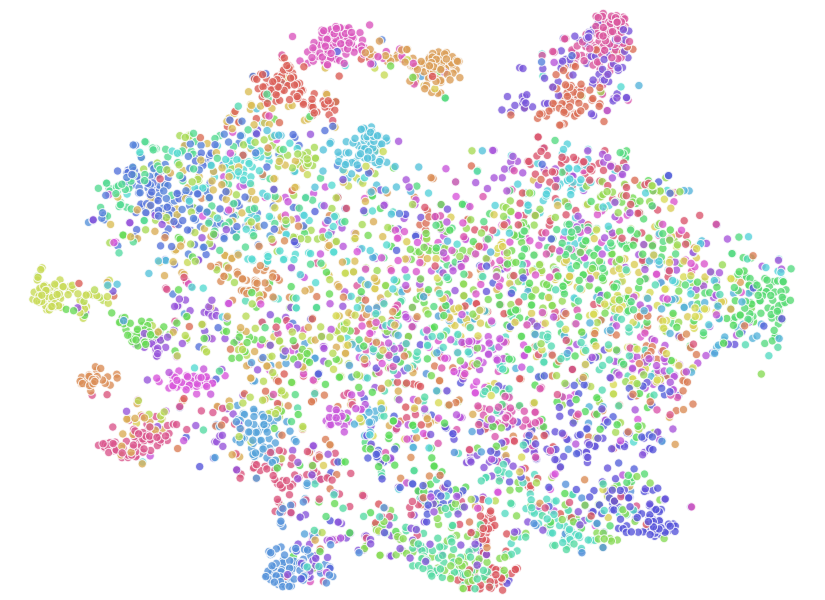}\hfill
    \includegraphics[width=.45\textwidth, height=0.2\textheight]{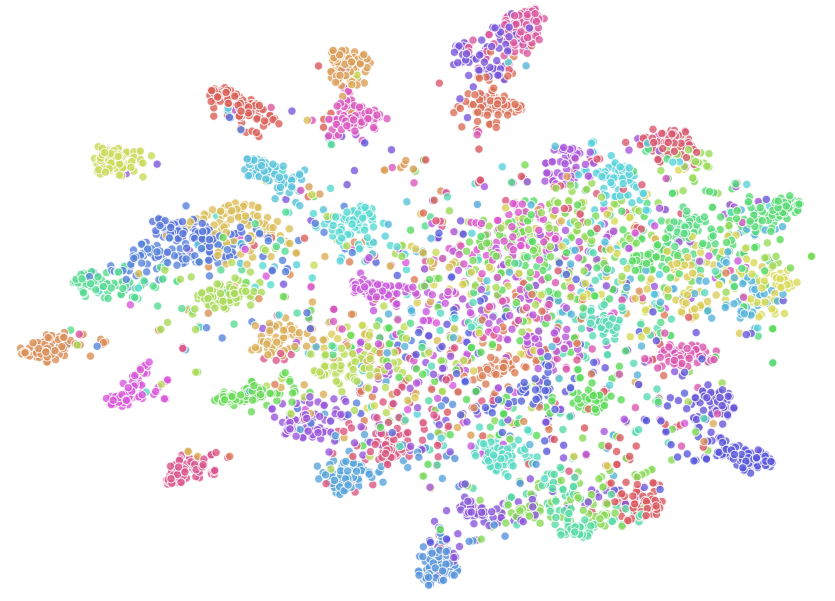}\hfill
    \caption{T-SNE \cite{JMLR:v9:vandermaaten08a} visualizations of the base $50$ classes of CIFAR100 in the embedding space, after all $100$ classes have been learned (\textbf{Left}: LUCIR \cite{Hou2019LearningAU}, \textbf{Right}: LUCIR+CSC)}
    \label{fig:cluster1}
\end{figure}

In Fig.~\ref{fig:cluster1}, we present T-SNE \cite{JMLR:v9:vandermaaten08a} visualizations of the embedding space, without and with our Cross-Space Clustering (CSC) objective (\ref{eqn:obj1}). The $50$ classes learned in the initial task are plotted in the embedding spaces of both models, once all the $100$ classes have been learned. 
It is seen that the CSC objective results in better clusters of prior classes in the feature space, compared to the baseline. The number of overlapping classes are reduced to a significant extent, as our objective ensures that the clusters are well-separated.

\section{Conclusion}


We introduced two complementary distillation-based objectives for class incremental learning. 
Our first objective called \textit{cross-space clustering}  positions classes appropriately in the embedding space and enables classes to counteract forgetting jointly.
Our second objective called \textit{controlled transfer} controls the positive and negative transfer between classes by estimating and utilizing inter-class relationships. We perform extensive experiments across a wide range of experimental settings to showcase the effectiveness of our objectives.

\noindent \textbf{Acknowledgements:} We are grateful to the Department of Science and Technology, India, as well as Intel India for the financial support of this project through the IMPRINT program (IMP/2019/000250) as well as the DST ICPS Data Science Cluster program. KJJ thanks TCS for their PhD Fellowship. We also thank the anonymous reviewers and Area Chairs for their valuable feedback in improving the presentation of this paper.

%
%
\bibliographystyle{splncs04}
\bibliography{egbib}

\newpage
\pagebreak
\definecolor{Gray}{gray}{0.95}
\definecolor{OliveGreen}{cmyk}{0.64,0,0.95,0.40}
\definecolor{ao(english)}{rgb}{0.0, 0.5, 0.0}
\definecolor{brightmaroon}{rgb}{0.76, 0.13, 0.28}

\addtolength{\textfloatsep}{-0.2in} 

\setcounter{table}{0}
\renewcommand{\thetable}{S\arabic{table}}

\renewcommand{\thefigure}{A\arabic{figure}}
\setcounter{figure}{0}

\begin{center}
\textbf{\Large Class-Incremental Learning with Cross-Space Clustering and Controlled Transfer: Supplementary Material}
\end{center}

\section{Results on Different Batch Sizes}

Our method leverages examples from the entire batch in order to facilitate training. All our results on CIFAR100 in the original paper are reported using a batch size of $128$, following prior work \cite{Hou2019LearningAU, Douillard2020PODNetPO}. Here, we analyze the efficacy of our method on different batch sizes used during training. Prior work has observed that increasing the batch size tends to slightly reduce the test accuracy, and have proposed various methods to reduce the drop in accuracy \cite{l.2018dont, goyal2017accurate, you2019large}. In our experiments, we adopt the linear scaling rule \cite{goyal2017accurate}, which scales the learning rate in proportion to the batch size. \\

\setlength\intextsep{0pt} 
\begin{wraptable}{r}{0.6\textwidth}
  {\scriptsize
  \caption{Ablation studies on the Batch Size}
  \label{table:batch_size}}
  \begin{center}
    \scriptsize
    \begin{tabular}{l|l|l|l}
        \toprule
        \multicolumn{2}{c|}{Settings} & 
        \multicolumn{1}{c|}{$\mathcal{B}=50$} & \multicolumn{1}{c}{$\mathcal{B}=\mathcal{C}$} 
        \\ \midrule
        \multicolumn{1}{c|}{Methods} & 
        \multicolumn{1}{c|}{Batch Size} & 
        \multicolumn{2}{c}{$\mathcal{C}=5$}   
        \\ \midrule
        \multicolumn{1}{l|}{LUCIR \cite{Hou2019LearningAU}} & 
        \multicolumn{1}{l|}{128} & 
        \multicolumn{1}{l|}{$59.4$} & 
        $42.28$ \\
        \rowcolor{Gray}
        \multicolumn{1}{l|}{LUCIR + CSCCT} & 
        \multicolumn{1}{l|}{} & 
        \multicolumn{1}{l|}{$60.01_{\color{red}\textbf{+2.61}}$} & 
        $44.03_{\color{red}\textbf{+1.55}}$ \\
        
        \multicolumn{1}{l|}{LUCIR \cite{Hou2019LearningAU}} & 
        \multicolumn{1}{l|}{256} & 
        \multicolumn{1}{l|}{$57.06$} & 
        $39.58$ \\
        \rowcolor{Gray}
        \multicolumn{1}{l|}{LUCIR + CSCCT} & 
        \multicolumn{1}{l|}{} & 
        \multicolumn{1}{l|}{$59.71_{\color{red}\textbf{+2.91}}$} & 
        $41.48_{\color{red}\textbf{+1.90}}$ \\
        
        \multicolumn{1}{l|}{LUCIR \cite{Hou2019LearningAU}} & 
        \multicolumn{1}{l|}{512} & 
        \multicolumn{1}{l|}{$56.19$} & 
        $38.16$ \\
        \rowcolor{Gray}
        \multicolumn{1}{l|}{LUCIR + CSCCT} & 
        \multicolumn{1}{l|}{} & 
        \multicolumn{1}{l|}{$59.02_{\color{red}\textbf{+2.83}}$} & 
        $40.2_{\color{red}\textbf{+2.04}}$ \\
        
        \multicolumn{1}{l|}{LUCIR \cite{Hou2019LearningAU}} & 
        \multicolumn{1}{l|}{1024} & 
        \multicolumn{1}{l|}{$54.83$} & 
        $37.97$ \\
        \rowcolor{Gray}
        \multicolumn{1}{l|}{LUCIR + CSCCT} & 
        \multicolumn{1}{l|}{} & 
        \multicolumn{1}{l|}{$57.8_{\color{red}\textbf{+2.97}}$} & 
        $40.27_{\color{red}\textbf{+2.3}}$ \\
        \bottomrule
    \end{tabular}%
  \end{center}
\end{wraptable}

Table \ref{table:batch_size} showcases the results on two different experimental settings, across various batch sizes. It can be seen that increasing the batch size increases the performance of our method. This is because larger batch sizes allow using more samples from the memory as well as the current task, providing a broader view of the feature space, which directly benefits our objectives.

\section{Results on Different Exemplar Memory Sizes}

The memory size specifies the \textit{exemplars-per-class} that the model can store at the end of each phase. Here, we report results varying the exemplar memory size. \newpage

\setlength\intextsep{0pt} 
\begin{wraptable}{r}{0.6\textwidth}
  {\scriptsize
  \caption{Ablation studies on the Memory Size}
  \label{table:exemplar_size}}
  \begin{center}
    \scriptsize
    \begin{tabular}{l|l|l|l}
        \toprule
        \multicolumn{2}{c|}{Settings} & 
        \multicolumn{1}{c|}{$\mathcal{B}=50$} & \multicolumn{1}{c}{$\mathcal{B}=\mathcal{C}$} 
        \\ \midrule
        \multicolumn{1}{c|}{Methods} & 
        \multicolumn{1}{c|}{Memory Size} & 
        \multicolumn{2}{c}{$\mathcal{C}=5$}   
        \\ \midrule
        \multicolumn{1}{l|}{LUCIR \cite{Hou2019LearningAU}} & 
        \multicolumn{1}{l|}{10} & 
        \multicolumn{1}{l|}{$55.83$} & 
        $39.45$ \\
        \rowcolor{Gray}
        \multicolumn{1}{l|}{LUCIR + CSCCT} & 
        \multicolumn{1}{l|}{} & 
        \multicolumn{1}{l|}{$57.54_{\color{red}\textbf{+1.71}}$} & 
        $40.72_{\color{red}\textbf{+1.14}}$ \\
        
        \multicolumn{1}{l|}{LUCIR \cite{Hou2019LearningAU}} & 
        \multicolumn{1}{l|}{20} & 
        \multicolumn{1}{l|}{$59.4$} & 
        $42.28$ \\
        \rowcolor{Gray}
        \multicolumn{1}{l|}{LUCIR + CSCCT} & 
        \multicolumn{1}{l|}{} & 
        \multicolumn{1}{l|}{$60.01_{\color{red}\textbf{+2.61}}$} & 
        $44.03_{\color{red}\textbf{+1.55}}$ \\
        
        \multicolumn{1}{l|}{LUCIR \cite{Hou2019LearningAU}} & 
        \multicolumn{1}{l|}{30} & 
        \multicolumn{1}{l|}{$62.52$} & 
        $46.35$ \\
        \rowcolor{Gray}
        \multicolumn{1}{l|}{LUCIR + CSCCT} & 
        \multicolumn{1}{l|}{} & 
        \multicolumn{1}{l|}{$65.05_{\color{red}\textbf{+2.53}}$} & 
        $48.34_{\color{red}\textbf{+1.99}}$ \\
        
        \multicolumn{1}{l|}{LUCIR \cite{Hou2019LearningAU}} & 
        \multicolumn{1}{l|}{40} & 
        \multicolumn{1}{l|}{$63.60$} & 
        $50.16$ \\
        \rowcolor{Gray}
        \multicolumn{1}{l|}{LUCIR + CSCCT} & 
        \multicolumn{1}{l|}{} & 
        \multicolumn{1}{l|}{$66.49_{\color{red}\textbf{+2.89}}$} & 
        $52.28_{\color{red}\textbf{+2.12}}$ \\
        \bottomrule
    \end{tabular}%
  \end{center}
\end{wraptable}

\noindent  Table \ref{table:exemplar_size} showcases the results. Enforcing stricter memory constraints causes a performance drop in LUCIR \cite{Hou2019LearningAU}, however, our method still provides strong relative improvements across settings. As the memory size increases, our method offers greater relative improvements.

\section{Phase-Wise Plots}

\begin{figure}[h]
    \centering
    \includegraphics[width=1\textwidth]{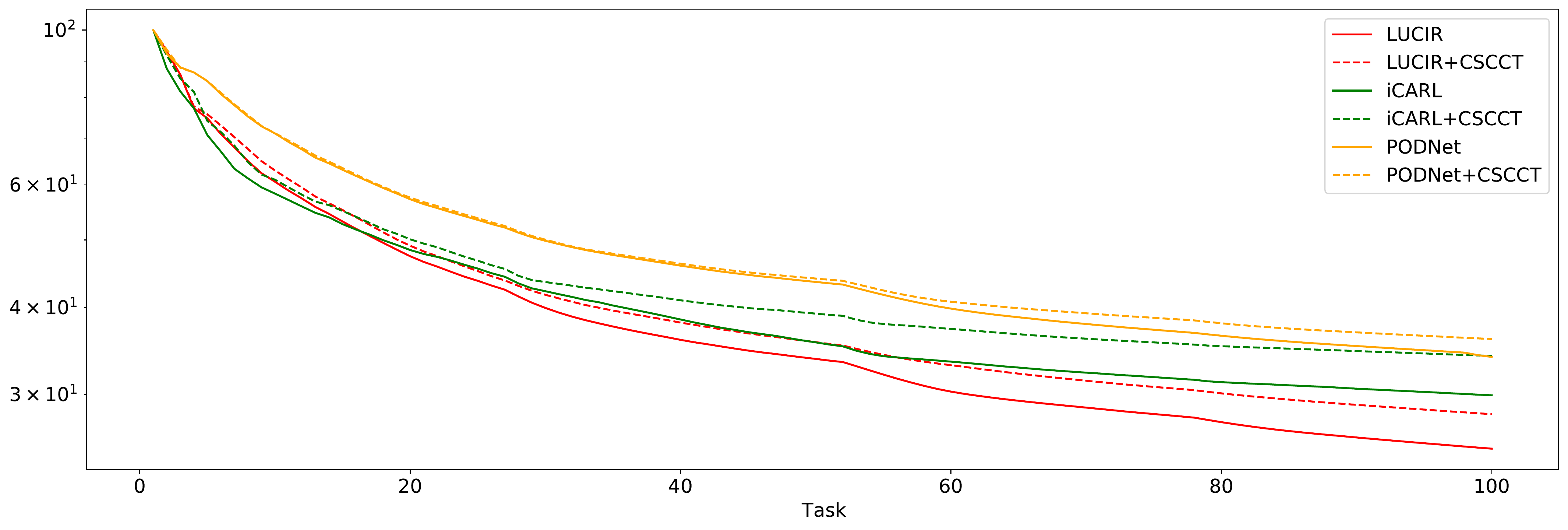}\hfill
    \caption{Phase-wise average incremental accuracies on CIFAR100, on the $100$ Task Setting with $1$ Class per Task. The y-axis is set to log scale for visual clarity.}
    \label{fig:1_1}
\end{figure}

\begin{figure}[h]
    \centering
    \includegraphics[width=0.57\textwidth]{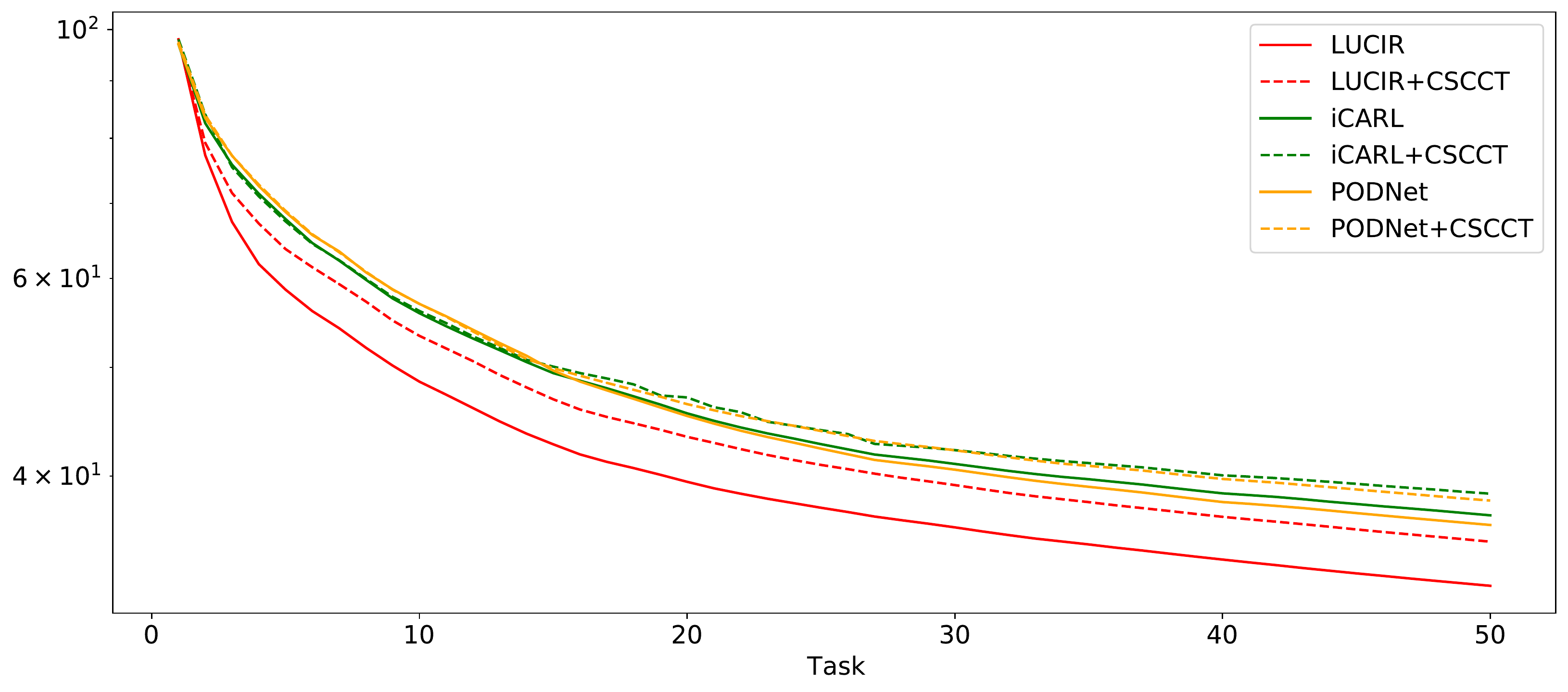}\hfill
    \includegraphics[width=.43\textwidth]{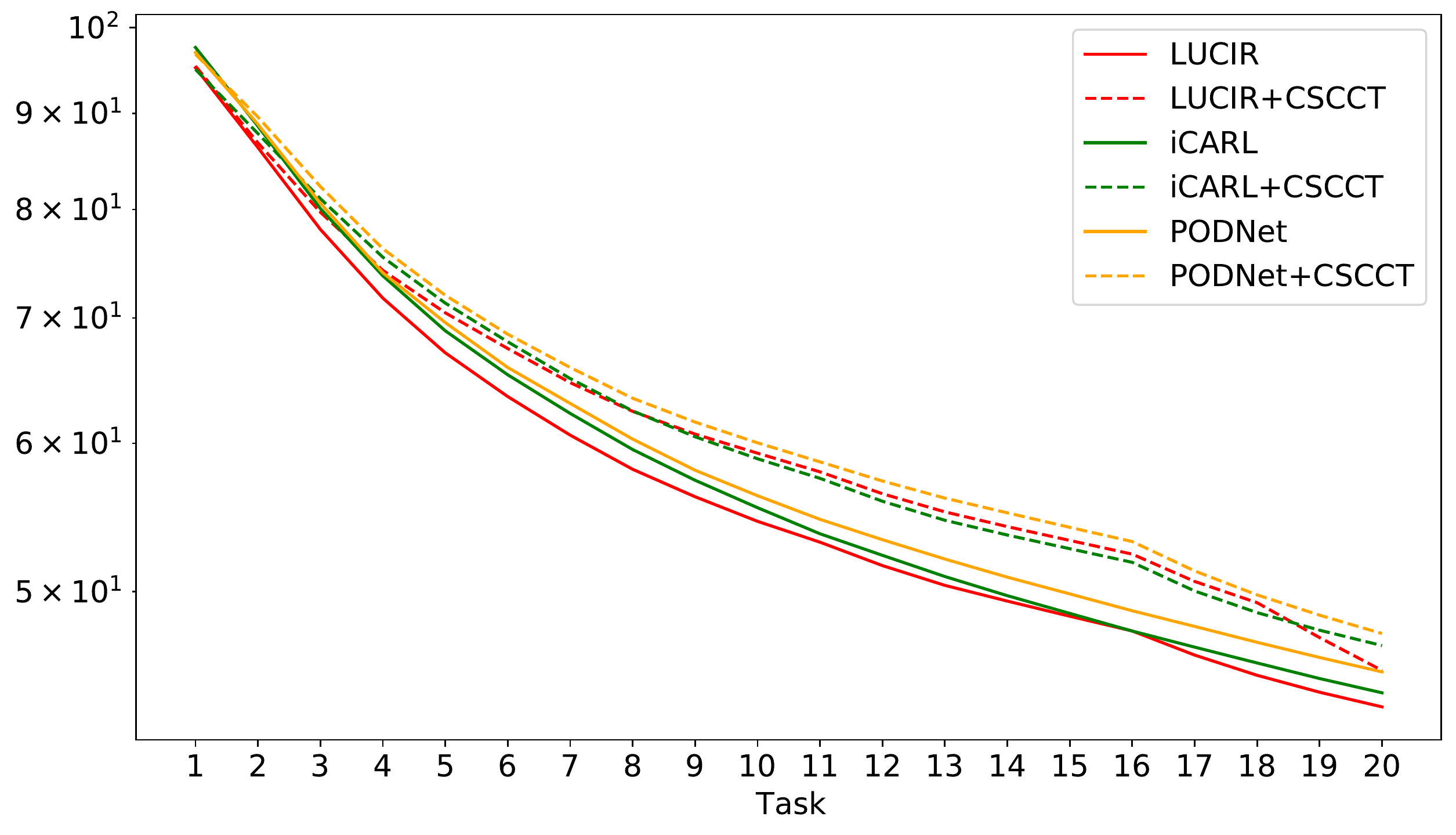}\hfill
    \caption{Phase-wise average incremental accuracies on CIFAR100, on the $50$ Task Setting with $2$ classes per task (Left) and the $20$ Task Setting with $5$ classes per task (Right). The y-axis is set to log scale for visual clarity.}
    \label{fig:2_2_5_5}
\end{figure}

Figures \ref{fig:1_1} and \ref{fig:2_2_5_5} showcase phase-wise plots on three class-incremental learning settings on CIFAR100, on top of multiple baseline methods.

%
%

\end{document}